%% file: main.tex
\documentclass[10pt,twocolumn,letterpaper]{article}

\usepackage{cvpr}              

\usepackage[accsupp]{axessibility}  

\input{preamble}

\definecolor{cvprblue}{rgb}{0.21,0.49,0.74}
\usepackage[pagebackref,breaklinks,colorlinks,allcolors=cvprblue]{hyperref}
\usepackage[capitalize]{cleveref}

\newcommand\blfootnote[1]{%
  \begingroup
  \renewcommand\thefootnote{}\footnote{#1}%
  \addtocounter{footnote}{-1}%
  \endgroup
}

\newif\ifdraft
 \draftfalse
\ifdraft
\newcommand{\jwc}[1]{{\color{orange}[\textbf{Jian} #1]}}
\newcommand{\smc}[1]{{\color{red}[\textbf{Sizhuo} #1]}}

\else
\newcommand{\jwc}[1]{}
\newcommand{\smc}[1]{}

\fi

\def\paperID{35} 
\def\confName{CVPR}
\def\confYear{2026}

\title{Bounded-Compute Multimodal Regression for Product-Rating Prediction}

\author{
William Leach \quad Ru He \quad Sizhuo Ma \quad Yizhen Jia \quad Min Cao \quad Jian Wang \quad Rick Cao \quad\\
Snap Inc.\\
{\tt\small wleach@snapchat.com}
}

\begin{document}
\maketitle
\input{sec/0_abstract}
\blfootnote{\copyright~2026 IEEE. Personal use of this material is permitted. Permission from IEEE must be obtained for all other uses, in any current or future media, including reprinting/republishing this material for advertising or promotional purposes, creating new collective works, for resale or redistribution to servers or lists, or reuse of any copyrighted component of this work in other works.}
\input{sec/1_intro}
\input{sec/2_related_work}
\input{sec/3_method}
\input{sec/4_experiments}
\input{sec/5_conclusion}

{
    \small
    \bibliographystyle{ieeenat_fullname}
    \bibliography{main}
}


\end{document}

%% file: preamble.tex
\usepackage{graphicx}
\usepackage{amsmath}
\usepackage{amssymb}
\usepackage{booktabs}
\usepackage[shortlabels]{enumitem}



%% file: sec/0_abstract.tex
\begin{abstract}
Vision-language models (VLMs) are increasingly attractive for multimodal quality assessment, but their default reliance on autoregressive text generation and dynamic visual processing is poorly matched to scalar regression under strict latency budgets. We present a bounded-compute adaptation of SmolVLM2-256M-Video-Instruct 
\cite{marafioti2025smolvlm}
for product-rating prediction in the LoViF 2026 Efficient VLM challenge. Motivated by recent multimodal engagement-prediction results showing that feature-based regression can outperform token-based score generation \cite{sun2025engagement}, we replace the language-modeling head with a lightweight two-layer MLP fed by pooled decoder states, and we enforce deterministic inputs through fixed $384\times384$ images and truncated metadata. Across controlled ablations, static global image processing slightly outperforms dynamic tiling, and scaling from 100K to 16M training examples substantially improves validation correlation. Under the official held-out evaluation, our 228M-parameter model achieves 0.39 PLCC and 0.40 CES, providing a strong and reproducible baseline for resource-constrained multimodal regression.
\end{abstract}

%% file: sec/1_intro.tex
\section{Introduction}
\label{sec:intro}

Vision-language models (VLMs) have become a standard foundation for multimodal reasoning, with strong performance on visual question answering, captioning, document understanding, and image-text dialogue \cite{alayrac2022flamingo,li2023blip2,liu2024llava15,laurencon2024buildingvlm}. However, most high-performing VLMs are optimized for open-ended text generation rather than bounded-cost scalar prediction. This mismatch is especially problematic in deployment settings that require predictable latency, fixed memory usage, and high throughput.

We address this bounded-compute regression challenge through the lens of \emph{automated product-rating prediction} using product images and structured metadata. Although this task only requires a single scalar output, a standard generative VLM pipeline still incurs autoregressive decoding overhead and often relies on dynamic image tiling or resizing to preserve fine-grained visual detail \cite{laurencon2024idefics2,Bai2025Qwen25VLTR}. These design choices are sensible for open-ended recognition and reasoning, but they produce variable sequence lengths and input-dependent compute, which is undesirable for large-scale or real-time regression systems.

This problem is formalized by the LoViF 2026 Challenge on Efficient VLM for Multimodal Creative Quality Scoring, which evaluates product-rating prediction under joint accuracy and efficiency constraints \cite{lovif2026challenge}. Recent work on short-video engagement prediction \cite{li2024delving,li2025vquala} provides an important clue for this setting: Sun et al.~\cite{sun2025engagement} show that feature-based regression on top of multimodal hidden states outperforms token-based score generation for Qwen2.5-VL, suggesting that hidden-state regression is often a better fit than autoregressive decoding when the target is a continuous score.

Motivated by this observation, we adapt SmolVLM2-256M-Video-Instruct into a deterministic multimodal regressor \cite{marafioti2025smolvlm}. We replace the language-modeling head with a lightweight two-layer MLP, pool the decoder hidden states into a fixed representation, and enforce bounded compute by resizing all images to a fixed resolution and aggressively truncating metadata. Our backbone choice is deliberate: SmolVLM2 is a compact VLM family built for efficient multimodal inference, and the 256M Video-Instruct checkpoint combines a SigLIP vision encoder with a SmolLM2 decoder in an Idefics3-style design \cite{marafioti2025smolvlm,smolvlm2modelcard,laurencon2024buildingvlm}.

\begin{figure*}[t]
  \centering
  \includegraphics[width=0.9\textwidth]{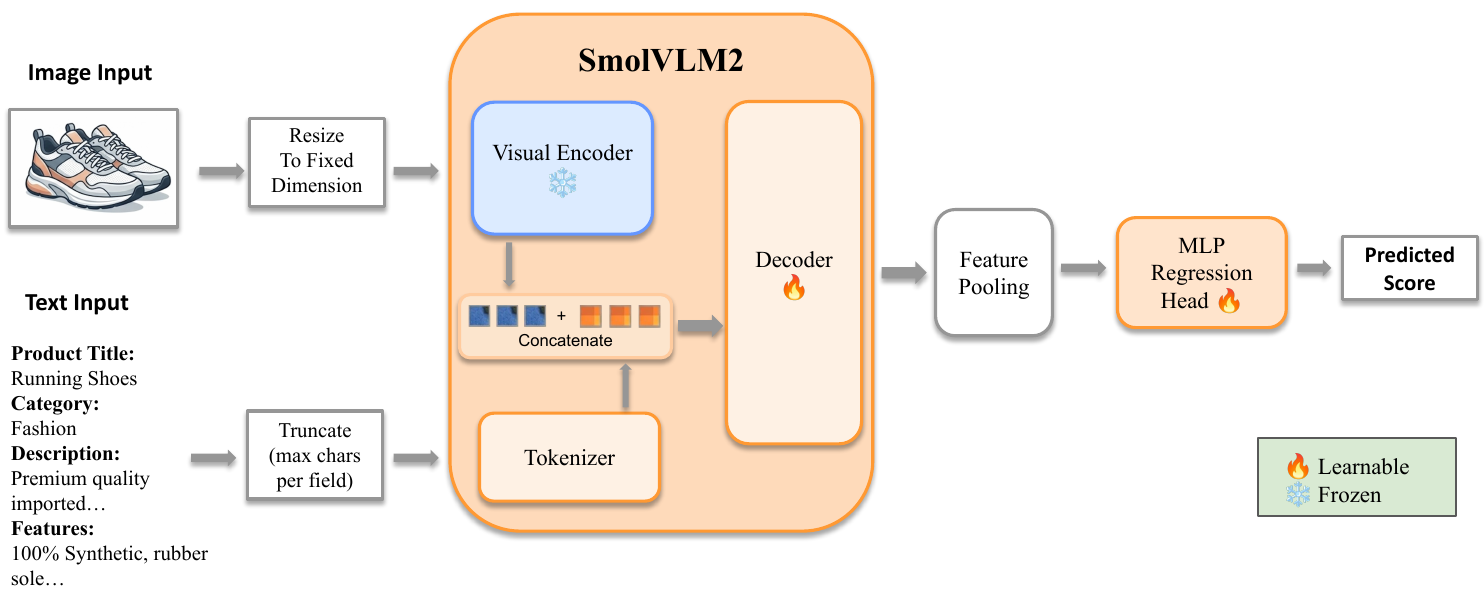}
  \caption{Overview of our bounded-compute regression pipeline. A fixed-resolution image and truncated text metadata are encoded into a multimodal token sequence, processed by SmolVLM2, pooled into a single representation, and mapped to a scalar rating by a lightweight MLP head.}
  \label{fig:pipeline}
\end{figure*}

Our contributions are threefold. First, we present a simple but effective recipe for converting a compact generative VLM into a bounded-compute multimodal regressor. Second, through controlled ablations, we show that static global image processing can slightly outperform dynamic tiling for this global-context scoring task, while short metadata truncation provides a favorable efficiency--accuracy trade-off. Third, we demonstrate that a 256M-class VLM continues to benefit from large-scale supervision, scaling effectively from 100K to 16M Amazon Reviews'23 examples while maintaining strong performance under the official LoViF evaluation protocol.

%% file: sec/2_related_work.tex
\section{Related Work}

\subsection{Vision-Language Models}
Representative modern vision-language model (VLM) families include Flamingo \cite{alayrac2022flamingo}, BLIP-2 \cite{li2023blip2}, LLaVA-1.5 \cite{liu2024llava15}, and Idefics3-style architectures \cite{laurencon2024buildingvlm}. The Qwen family has significantly influenced the field: Qwen-VL established multilingual grounding and OCR \cite{qwen2023qwenvl}, while Qwen2-VL and Qwen2.5-VL introduced dynamic resolution and enhanced document/video parsing \cite{qwen2024qwen2vl,Bai2025Qwen25VLTR}. More recently, the evolution of the backbone family from Qwen2 to Qwen3 has further pushed the boundaries of multimodal performance \cite{qwen2024qwen2vl,qwen2025qwen3}.

However, these models primarily focus on open-ended text generation, which introduces variable latency and input-dependent compute \cite{laurencon2024idefics2,Bai2025Qwen25VLTR}. 
A parallel line of research focuses on compact and efficient VLMs for constrained deployment, such as MobileVLM \cite{chu2023mobilevlm}, MobileVLM V2 \cite{chu2024mobilevlmv2}, MiniCPM-V \cite{yao2024minicpmv}, PaliGemma 2 \cite{steiner2024paligemma2}, and DeepSeek-VL2 \cite{wu2024deepseekvl2}. 
\citet{marafioti2025smolvlm} introduced SmolVLM, aimed at resource-efficient inference, 
by systematically studying the factors that affect compact VLMs,
including the balance between vision encoder and language-model decoder capacity,
tokenization, positional encoding, text prompt engineering, and training-data composition.
They subsequently released SmolVLM2, a family of VLMs (ranging from 256M to 2.2B parameters)
with even better visual understanding capabilities \cite{smolvlm2modelcard}. 

While our work aligns with this compact-VLM trajectory, we differentiate ourselves by focusing on \textbf{bounded-compute scalar regression} rather than generative tasks.

\subsection{Multimodal Regression and Quality Scoring}
Recent research has increasingly focused on adapting large multimodal models for scalar prediction and perceptual assessment. For instance, DepictQA explores multimodal language models for image quality assessment \cite{you2023depictqa}, while AesBench provides a benchmark for evaluating image aesthetics perception in large models \cite{huang2024aesbench}. 
In the domain of short-video engagement, Li et al. \cite{li2024delving} introduce a large-scale benchmark dataset
comprising real-world user-generated content (UGC) short videos, and proposed a predictive model that uses a cross-modal attention mechanism to incorporate rich multi-modal features.

Critically, Sun et al. \cite{sun2025engagement} evaluated both feature-based regression and token-based score generation for engagement prediction on UGC using large multimodal models.
Their findings indicate that regression from hidden features via a lightweight MLP consistently outperforms autoregressive token generation when the target is a continuous score \cite{sun2025engagement}. 
This empirical evidence directly motivates our decision to repurpose a compact VLM into a deterministic regressor, replacing the language-modeling head with a regression-specific architecture to ensure both higher accuracy and stable computational overhead \cite{sun2025engagement, marafioti2025smolvlm}.

\subsection{Multimodal E-commerce Prediction and Recommendation}
Our task is closely related to multimodal recommendation and e-commerce modeling, where product images and metadata provide complementary signals regarding item quality and consumer preference. Early approaches like VBPR \cite{he2016vbpr} pioneered the integration of visual item features into recommendation frameworks. More recently, multimodal review analysis has explored predicting review helpfulness by combining product, review, image, and text signals through multi-perspective coherent reasoning \cite{liu2021mcr} or selective-attention contrastive learning \cite{han2022sancl}. 

As noted in recent surveys, multimodal item representations are increasingly critical for predictive tasks across online platforms~\cite{liu2023mmrecsurvey}. Our setting differs from these lines in its focus: we predict a product-level scalar rating directly from a primary image and structured metadata under strict parameter and FLOP budgets.

%% file: sec/3_method.tex
\section{Method Details}
\label{sec:method}

Our goal is to adapt a compact generative VLM into a deterministic regressor for product-rating prediction. The input to our model is a single product image paired with structured metadata fields (title, description, features, and main category), and the target is the product's average star rating in $[1,5]$.

\subsection{Backbone Choice}
We build upon SmolVLM2-256M-Video-Instruct, a compact instruction-tuned VLM that combines a SigLIP vision encoder with a SmolLM2 decoder in an Idefics3-style multimodal architecture \cite{marafioti2025smolvlm,smolvlm2modelcard,laurencon2024buildingvlm}. This model is particularly well-suited for the LoViF efficiency challenge because it retains strong multimodal grounding capabilities while operating at a much smaller scale than mainstream VLMs. Our adapted regression model contains 228M parameters in total, with 135M  trainable during fine-tuning.

\subsection{Deterministic Multimodal Processing}
A key design goal of our model is \emph{bounded compute}. Standard VLM pipelines often use dynamic image tiling or native dynamic resolution to preserve fine detail \cite{laurencon2024idefics2,Bai2025Qwen25VLTR}. While beneficial for OCR-heavy or fine-grained recognition tasks, those mechanisms introduce input-dependent visual token counts and therefore variable latency and memory usage.

To avoid this variability, we disable dynamic resizing and image splitting, and resize every image to a fixed $384\times384$ resolution using bilinear interpolation. On the text side, we concatenate the product's metadata fields into a structured natural language prompt. To guarantee a strictly bounded language context, each formatted key--value pair is individually truncated to a maximum character budget $L$ (e.g., $L=100$) before concatenation. The final multimodal prompt $P$ follows this fixed template:

\begin{quote}
\small \ttfamily
<image> The average user rating for this product. Text metadata: Title: <title>[:L], Description: <description>[:L], Features: <features>[:L], Main Category: <category>[:L]
\end{quote}

The entire prompt $P$ is tokenized and processed by the SmolVLM2 decoder. Unlike generative architectures that output numerical scores via autoregressive next-token generation (e.g., Qwen2.5-VL), our model aggregates the hidden states of this deterministic sequence and projects them directly into a continuous scalar rating. This yields strictly bounded multimodal sequence lengths, stable FLOPs, and highly reproducible runtime characteristics across all samples.

\subsection{Regression Adaptation}
While SmolVLM2 processes the visual and textual tokens with an autoregressive decoder, our task is constrained to a single scalar output. Following recent evidence that feature-based regression can outperform token-based score generation for multimodal engagement prediction \cite{sun2025engagement}, we remove the language-modeling head and attach a lightweight two-layer MLP regression head.

Given an RGB image $x$ and tokenized metadata, the SigLIP vision encoder first produces patch-level visual embeddings. These embeddings are compressed by the model's pixel-shuffle connector and projected into the decoder embedding space ($d=576$), then concatenated with text tokens and processed by the SmolLM2 decoder. Let $T$ denote the maximum sequence length in a batch, and $h_i \in \mathbb{R}^d$ the final hidden state at position $i$. To aggregate the variable-length decoder sequence into a fixed-dimensional representation, we employ mask-aware mean pooling:

\begin{equation}
\label{eq:mask_pool}
h_{\text{pool}} = \frac{\sum_{i=1}^{T} m_i h_i}{\sum_{i=1}^{T} m_i}.
\end{equation}

Here, $m_i \in \{0,1\}$ represents the standard binary attention mask generated natively by the multimodal processor during batching, $m_i = 1$ if the token at position $i$ is a valid visual or textual token, and $m_i = 0$ if the token at position $i$ is a padding token.

This formulation ensures that only meaningful multimodal features contribute to the final pooled state, explicitly preventing padding tokens from diluting the representation.

The pooled vector is then passed through a two-layer MLP head,
\texttt{Linear(576, 288) $\rightarrow$ ReLU $\rightarrow$ Linear(288, 1)},
which outputs a scalar logit $x$. To restrict predictions to the valid rating range, we apply a scaled sigmoid:

\begin{equation}
\label{eq:scaled_sigmoid}
\hat{y} = 1 + 4\,\sigma(x),
\end{equation}

where $\sigma(\cdot)$ is the logistic sigmoid function. This bounded parameterization explicitly constrains the output to the valid $[1, 5]$ interval, ensuring that the model avoids implausible out-of-range predictions by construction.

\subsection{Training Objective}
We optimize the model with mean squared error between predicted and ground-truth ratings:

\begin{equation}
\label{eq:mse}
L = \frac{1}{N} \sum_{i=1}^{N} (y_{\text{pred}}^{(i)} - y_{\text{true}}^{(i)})^2.
\end{equation}

During fine-tuning, the SigLIP vision encoder and pixel-shuffle connector are frozen, while the decoder and regression head are updated. This choice preserves the compact model's pretrained visual representation while focusing optimization on the multimodal decoder and the task-specific scalar head.

%% file: sec/4_experiments.tex
\section{Experiments}
\label{sec:experiments}

\subsection{Dataset and Evaluation Metrics}
We train our model on Amazon Reviews'23 \cite{hou2024bridging,amazonreviews23hf}, using item-level metadata paired with a representative product image. Each sample contains one primary product image together with four metadata fields: title, description, features, and main category. The regression target is the product's average star rating.

\paragraph{Category balancing.}
Amazon Reviews'23 is highly imbalanced across product categories: large groups such as \texttt{Unknown}, \texttt{Clothing\_Shoes\_and\_Jewelry}, and \texttt{Books} contain far more items than long-tail categories such as \texttt{Subscription\_Boxes}, \texttt{Gift\_Cards}, and \texttt{Magazine\_Subscriptions} \cite{amazonreviews23hf}. To reduce category dominance, we apply popularity-stratified sampling within each category: we select the top-1M and bottom-1M items ranked by \texttt{rating\_number}, and retain all items for categories with fewer than 2M candidates.

\paragraph{Filtering and preprocessing.}
We further filter the sampled items to those with (i) at least 10 reviews and (ii) an available \texttt{MAIN} image, preferring high-resolution URLs when available. This yields 16,455,671 training samples, with 10,000 held out for validation. All images are resized to $384\times384$ using bilinear interpolation; dynamic resizing and image splitting are disabled. Metadata fields are formatted as key--value text and truncated to 100 characters per field unless otherwise noted.

To evaluate predictive accuracy and ranking quality, we report:
\begin{itemize}[leftmargin=*]
    \item \textbf{RMSE (Root Mean Square Error):} average magnitude of the prediction error; lower is better.
    \item \textbf{PLCC (Pearson Linear Correlation Coefficient):} linear correlation between predicted and target ratings; higher is better.
    \item \textbf{SRCC (Spearman's Rank Correlation Coefficient):} rank correlation between predictions and targets; higher is better.
\end{itemize}

\subsection{Training and Implementation Details}
Unless otherwise noted, all experiments use SmolVLM2-256M-Video-Instruct with the deterministic $384\times384$ preprocessing pipeline described in \cref{sec:method}. The SigLIP vision encoder and pixel-shuffle connector are frozen, while the decoder and regression head (with 135M parameters)
are fine-tuned with MSE loss. Training is performed with DistributedDataParallel on 4 NVIDIA A100 GPUs using a global batch size of 64. We use 8-bit AdamW optimizer states \cite{dettmers20218}, a peak learning rate of $4\times10^{-4}$, linear decay, 3\% warmup, and 0.1 dropout in the regression head. We train for up to 5 epochs, employing early stopping to select the optimal model checkpoint based on peak validation performance.

At inference time, the adapted model contains 228M active parameters. On a single NVIDIA A100 with batch size 64, the final model runs at 0.0084 seconds per image (119.3 images/s). We also use FlashAttention-2 \cite{dao2023flashattention} and bfloat16 automatic mixed precision.

\subsection{Ablation Studies}
We perform controlled ablations to characterize the accuracy--efficiency trade-offs of compact multimodal regression. Unless otherwise stated, these ablations are conducted on a 1M-sample subset of the training data.

\paragraph{Model capacity.}
We first compare the 256M and 500M SmolVLM2 variants under the original dynamic $512\times512$ preprocessing pipeline. Table~\ref{tab:model_size} illustrates the predictive performance of both backbones.

\begin{table}[h]
\centering
\caption{Effect of model capacity (1M samples).}
\label{tab:model_size}
\begin{tabular}{lccc}
\toprule
\textbf{Backbone} & \textbf{RMSE} $\downarrow$ & \textbf{PLCC} $\uparrow$ & \textbf{SRCC} $\uparrow$ \\
\midrule
SmolVLM2-256M & 0.333 & 0.683 & 0.652 \\
SmolVLM2-500M & \textbf{0.329} & \textbf{0.694} & \textbf{0.664} \\
\bottomrule
\end{tabular}
\end{table}

While the 500M model yields better predictive correlation, the gain comes at a substantial resource cost. Specifically, upgrading to the 500M variant roughly doubles the active parameter count (460M vs. 228M) and raises the theoretical maximum compute from 113 GFLOPs to 165 GFLOPs. Given the strict efficiency constraints of our target application, the $+0.011$ gain in PLCC does not mathematically justify the increased memory footprint and computational overhead. We therefore retain the 256M model for all subsequent experiments to prioritize inference efficiency.

\paragraph{Static vs. dynamic visual preprocessing.}
The LoViF benchmark's canonical inference protocol strictly requires deterministic compute. Modern dynamic tiling generates a variable number of visual tokens depending on the input image's aspect ratio, resulting in heavily fluctuating, image-dependent FLOPs that cannot be objectively evaluated under the challenge's fixed-cost scoring system. To meet the benchmark requirements, we ablated the default dynamic tiling strategy against a forced static global resize at the same nominal $512\times512$ resolution. Because dynamic tiling compute varies per sample, we do not report a static FLOP comparison for this ablation.

\begin{table}[h]
\centering
\caption{Dynamic vs. static image preprocessing at $512\times512$.}
\label{tab:preprocessing}
\begin{tabular}{lccc}
\toprule
\textbf{Preprocessing} & \textbf{RMSE} $\downarrow$ & \textbf{PLCC} $\uparrow$ & \textbf{SRCC} $\uparrow$ \\
\midrule
Dynamic tiling & 0.333 & 0.683 & 0.652 \\
Static global resize & \textbf{0.331} & \textbf{0.689} & \textbf{0.657} \\
\bottomrule
\end{tabular}
\end{table}

Surprisingly, enforcing this strict operational constraint actually improved predictive performance. Despite using a theoretically less adaptive visual pipeline, static preprocessing proved slightly better in this setting. One possible explanation is that global product-rating prediction depends more on overall object presentation and coarse semantics than on localized high-resolution crops, making dynamic tiling unnecessary or even mildly distracting.

\paragraph{Input resolution.}
Having selected static preprocessing, we ablate image resolution. Table~\ref{tab:resolution} illustrates how scaling the image size impacts predictive performance.

\begin{table}[h]
\centering
\caption{Effect of static image resolution (1M samples).}
\label{tab:resolution}
\begin{tabular}{lccc}
\toprule
\textbf{Resolution} & \textbf{RMSE} $\downarrow$ & \textbf{PLCC} $\uparrow$ & \textbf{SRCC} $\uparrow$ \\
\midrule
$512 \times 512$ & \textbf{0.331} & \textbf{0.689} & \textbf{0.657} \\
$384 \times 384$ & 0.335 & 0.679 & 0.646 \\
\bottomrule
\end{tabular}
\end{table}

While the higher $512\times512$ resolution yields a slight improvement in raw correlation, it also substantially increases the computational cost. Specifically, increasing the resolution from $384$ to $512$ raises the theoretical maximum compute from 72 GFLOPs to 113 GFLOPs. Given the strict operational constraints of the target benchmark, we find that this $\sim 57\%$ increase in FLOPs outweighs the $+0.010$ gain in PLCC. Consequently, we adopt $384\times384$ as our final operating point to maintain a highly efficient compute profile.

\paragraph{Text truncation.}
To bound language-side compute, we vary the metadata truncation limit. Table~\ref{tab:text_length} illustrates the impact of different character limits on predictive performance.

\begin{table}[h]
\centering
\caption{Effect of metadata truncation at $384\times384$.}
\label{tab:text_length}
\begin{tabular}{lccc}
\toprule
\textbf{Char limit} & \textbf{RMSE} $\downarrow$ & \textbf{PLCC} $\uparrow$ & \textbf{SRCC} $\uparrow$ \\
\midrule
50  & 0.340 & 0.666 & 0.628 \\
100 & 0.335 & 0.679 & 0.646 \\
200 & \textbf{0.333} & \textbf{0.685} & \textbf{0.648} \\
\bottomrule
\end{tabular}
\end{table}

While longer metadata naturally improves correlation, it introduces a substantial computational overhead. Specifically, expanding the truncation limit from 50 to 100 and 200 characters increases the theoretical maximum compute from 65 GFLOPs to 72 GFLOPs and 86 GFLOPs, respectively. Although the 200-character limit yields the highest overall PLCC, the additional $+0.006$ gain over the 100-character limit requires a disproportionate increase of 14 GFLOPs per forward pass. Balancing this trade-off, we select 100 characters per field as our highly efficient default operating point.

\subsection{Data Scaling and Final Results}
Using the final efficient configuration (SmolVLM2-256M, static $384\times384$ preprocessing, 100-character truncation), we evaluate how performance changes with training-set scale.

\begin{table}[h]
\centering
\caption{Effect of training data scale on validation performance.}
\label{tab:data_scale}
\begin{tabular}{lccc}
\toprule
\textbf{Training data} & \textbf{RMSE} $\downarrow$ & \textbf{PLCC} $\uparrow$ & \textbf{SRCC} $\uparrow$ \\
\midrule
$\sim$100K samples & 0.363 & 0.605 & 0.558 \\
$\sim$1M samples   & 0.335 & 0.679 & 0.646 \\
$\sim$16M samples  & \textbf{0.326} & \textbf{0.700} & \textbf{0.664} \\
\bottomrule
\end{tabular}
\end{table}

As shown in Table~\ref{tab:data_scale}, the compact backbone continues to benefit from additional supervision: scaling from 100K to 16M samples improves PLCC by +0.095 and SRCC by +0.106 on the same validation split. This is an important result for the challenge setting, since it shows that a small VLM can convert large external data into measurable quality gains without changing the deployment footprint.

\paragraph{Official challenge results.}
Following the LoViF challenge protocol \cite{lovif2026challenge}, the canonical evaluation heavily emphasizes operational cost via the Comprehensive Efficiency Score (CES):
$$ \text{CES} = \text{PLCC}^+ \times \mathcal{E}(\mathcal{C}) $$

Here, $\text{PLCC}^+ = \max(0, \text{PLCC}(y, \hat{y}))$ clips negatively correlated predictions to zero to avoid misaligned incentives, and $\mathcal{E}(\mathcal{C})$ is an asymmetric efficiency multiplier based on the geometric resource cost, $\mathcal{C}$. The cost $\mathcal{C}$ balances the model's parameters and compute against baseline targets ($P_{\text{tgt}} = 1000\text{M}$ parameters, $F_{\text{tgt}} = 20\text{G}$ FLOPs):
$$ \mathcal{C} = \left(\frac{\text{Params}}{P_{\text{tgt}}}\right)^{0.5} \cdot \left(\frac{\text{FLOPs}}{F_{\text{tgt}}}\right)^{0.5} $$

To encourage lightweight design, the efficiency factor $\mathcal{E}(\mathcal{C})$ applies a bounded bonus for efficient models ($\mathcal{C} \leq 1$) and a steep logarithmic penalty for over-budget models ($\mathcal{C} > 1$):
$$ \mathcal{E}(\mathcal{C}) =
\begin{cases}
\min(1 + 0.05 \ln(1/\mathcal{C}), 1.10) & \text{if } \mathcal{C} \leq 1 \\
\frac{1}{1 + 2.0 \ln(\mathcal{C})} & \text{if } \mathcal{C} > 1
\end{cases} $$

By calculating the average text length of the metadata fields on the final test set, we estimate an operational compute cost of 68 GFLOPs. Given our 228M parameter footprint, this yields a resource cost of $\mathcal{C} \approx 0.881$. Because this remains under the reference budget, we achieve a positive efficiency multiplier of $\mathcal{E}(\mathcal{C}) \approx 1.006$. Under the official held-out test evaluation, our final submission achieves \textbf{0.39 PLCC} and \textbf{0.40 CES}, \textbf{ranked 3rd on the leaderboard}. Because the official benchmark uses a separate hidden split and challenge-specific efficiency scoring, these results should be interpreted separately from the validation ablations above.

\subsection{Prediction--Target Alignment}
\begin{figure}[t]
  \centering
  \includegraphics[width=1.0\linewidth]{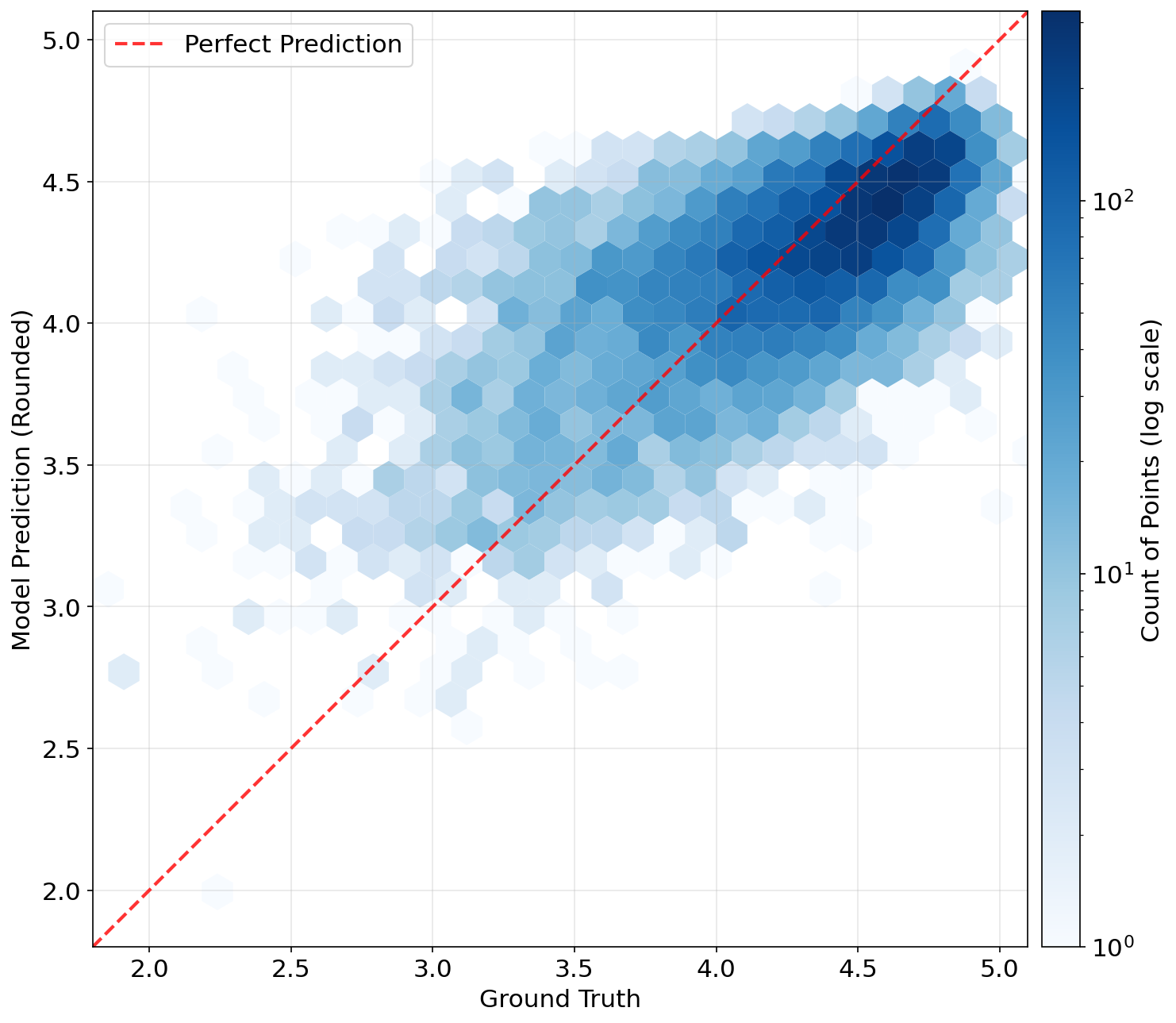}
  \caption{Hexbin density of rounded predictions versus rounded ground-truth ratings on the validation set. Predictions are rounded to one decimal place for visualization to match the label granularity. The dashed red line denotes perfect prediction.}
  \label{fig:hexbin}
\end{figure}

Figure~\ref{fig:hexbin} shows a strong concentration of mass near the diagonal, indicating that the model tracks the overall rating trend well. Errors are concentrated primarily in the mid-range ratings, where many products occupy visually similar quality bands and the target distribution is denser. The bounded sigmoid head also keeps all predictions within the valid $[1,5]$ interval by construction.

%% file: sec/5_conclusion.tex
\section{Conclusion}
\label{sec:conclusion}

We presented a bounded-compute adaptation of SmolVLM2 for multimodal product-rating prediction. By replacing autoregressive score generation with a lightweight regression head, and by enforcing fixed-resolution visual inputs and aggressively truncated metadata, we obtain a deterministic inference pipeline with stable latency and memory usage.

Our experiments show that strong scalar prediction does not require the full dynamic visual processing pipeline commonly used in open-ended VLMs. In particular, static global image processing slightly outperforms dynamic tiling in our setting, while compact text truncation preserves most of the accuracy at a much lower compute cost. We also find that the 256M backbone continues to benefit from large-scale supervision, improving steadily up to 16M training examples and achieving 0.39 PLCC / 0.40 CES on the official held-out evaluation.

These results position compact VLMs as a practical foundation for resource-constrained multimodal regression. At the same time, our design deliberately favors deterministic global processing over adaptivity, which may limit performance on products whose rating depends on fine-grained local details or long-form metadata. Future work could explore conditional compute allocation, stronger multimodal pretraining tailored to scoring tasks, or explicit uncertainty estimation for ambiguous examples.